\documentclass{llncs}

\usepackage{amsmath}
\usepackage{amsfonts}
\usepackage{booktabs}
\usepackage{caption}
\usepackage{subcaption}
\usepackage{graphicx}
\usepackage{pgfplots}
\usepackage[all]{nowidow}
\usepackage[utf8]{inputenc}
\usepackage{tikz}
\usetikzlibrary{er,positioning,bayesnet}
\usepackage{multicol}
\usepackage{algpseudocode,algorithm,algorithmicx}
\usepackage[inline]{enumitem}

\definecolor{blue}{HTML}{1F77B4}
\definecolor{orange}{HTML}{FF7F0E}
\definecolor{green}{HTML}{2CA02C}

\pgfplotsset{compat=1.14}

\begin{document}

\title{Deep Manifold Part 1: Anatomy of Neural Network Manifold}

\author{Max Y. Ma \thanks{corresponding author: paper@deepmanifold.ai}  and Gen-Hua Shi}

\institute{deepManifold}

\maketitle

\begin{abstract}
Based on the numerical manifold method principle, we developed a mathematical framework of a neural network manifold: Deep Manifold and discovered that neural networks: 1) is numerical computation combining forward and inverse; 2) have near infinite degrees of freedom; 3) exponential learning capacity with depth; 4) have self-progressing boundary conditions; 5) has training hidden bottleneck. We also define two concepts: neural network learning space and deep manifold space and introduce two concepts: neural network intrinsic pathway and fixed point. We raise three fundamental questions: 1). What is the training completion definition; 2). where is the deep learning convergence point (neural network fixed point); 3). How important is token timestamp in training data given negative time is critical in inverse problem.

\end{abstract}
\section{Introduction}

A neural network is a powerful numerical computation architecture. To evaluate the capacity of any numerical computation architecture, there are three fundamental questions to consider:

\begin{itemize}
    \item What degree of freedom does the system have? This will determine how many dimensions the system can compute.
    \item What degree of non-linearity can the system handle? This will determine how complex a problem the system is capable of solving.
    \item What kind of boundary conditions can the system handle? This will determine how well the system can manage external constraints.
\end{itemize}

We did not initially start with such a broad scope. Our initial effort was focused on understanding the nonlinear aspects of neural networks. Naturally and gradually, the scope expanded. We aimed to ensure our work was mathematically grounded, and the numerical manifold method became a natural choice.

\section{Numerical Manifold Method}
\begin{figure}[H]
    \centering
    \includegraphics[width=0.8\linewidth]{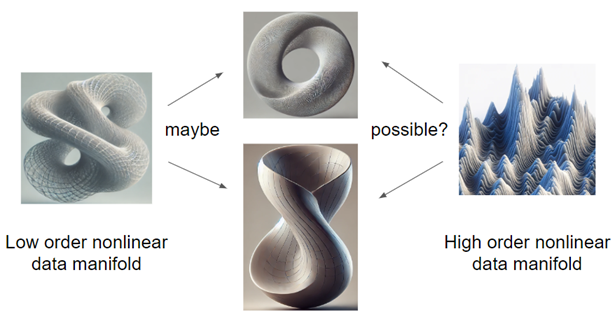}
    \caption{Data Manifold and Predefined Smooth Manifold }
    \label{fig:fig1}
\end{figure}

Manifolds are familiar to deep learning researchers and practitioners, especially for their ability in nonlinear dimensionality reduction. The principle behind these methods is to transform or map the data manifold to a predefined manifold space, such as a Möbius Strip or a Klein bottle (Figure \ref{fig:fig1}). For instance, a torus manifold can have up to 400 dimensions, but it remains on a smooth surface, which involves low-order non-linearity. When the training data manifold is smooth, this manifold transformation or mapping can achieve reasonable results due to minimal information loss during the transformation.

However, in the new era of GPT or AGI, where foundation models are trained with multi-trillion tokens, the training data manifolds are expected to exhibit extremely high dimensions and high-order non-linearity. The traditional manifold's ability for nonlinear dimensionality reduction appears to be reaching its limits.

The Numerical Manifold Method \cite{shi1991manifold} was developed with support from the US Department of Defense (DoD) in the early 1990s. One core principle of the Numerical Manifold Method is topology cover. See the illustration below of a single person's viewpoint/area for further explanation

\begin{figure}[H]
    \centering
    \includegraphics[width=0.4\linewidth]{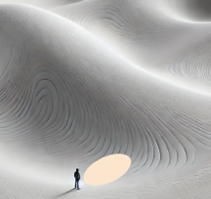}
    \caption{One Topology Cover}
    \label{fig:fig2}
\end{figure}

An integrable function can be applied to the cover to perform computations. Examples include the Fourier function (Ma, M.Y, 1995)\cite{ma1995single} and third-order functions (Ma, et al., 1996 )\cite{ma1996discontinuous}. Researchers have successfully developed a unified framework using the Numerical Manifold Method for various fields in the physical world, including solid dynamics, fluid dynamics, electrodynamics, thermodynamics, and acoustics.

The main reason for its success is the efficient computation it offers for high-order nonlinear problems. The high order nonlinear is defined below when $J > 4$

\begin{equation}
    f = \sum_{j=0}^{J}(\alpha_j x^j)
\end{equation}

There is no general closed-form solution using radicals when $J > 4$, but solutions can be found using numerical methods such as the numerical manifold method. See below illustration
\begin{figure}[H]
    \centering
    \includegraphics[width=0.75\linewidth]{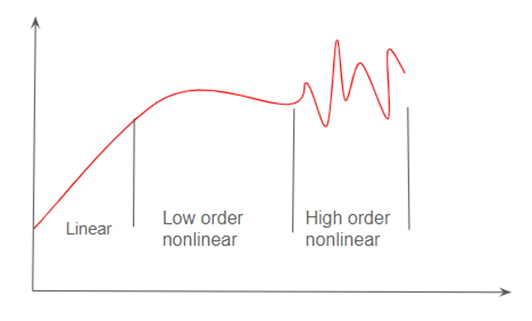}
    \caption{Linear and Nonlinear}
    \label{fig:fig3}
\end{figure}

There are two nonlinear properties within neural networks:
\begin{itemize}
    \item Local: nonlinear activation function \& dropout at node level.
    \item Globe: nonlinear transformation between hidden layers.
\end{itemize}

Significant deformation of feature maps has been observed, as reported by Duvenaud et al. (2015) \cite{duvenaud2016avoidingpathologiesdeepnetworks} and Ghorbani et al. (2018) \cite{ghorbani2018interpretationneuralnetworksfragile}, who noted that this deformation increases with neural network depth and training iterations. We interpret their findings as evidence of the neural network's high-order non-linearity.

In situations of high-order non-linearity, the derivative's operating range is very narrow, on the order of the inverse of $x$ to the power of $J$. It feels like sculpting sharp, rigid and fragile art glass, which requires a delicate touch and steady hands.

Besides its capability to handle high-order non-linearity, the Numerical Manifold Method can also solve both forward and inverse problems. We employ the principles of the Numerical Manifold Method to study the anatomy of neural network manifolds.

\section{Anatomy of Neural Network Manifold}
\subsection{Inverse and Forward Enablement}
\label{inverse_forward}
Neural network is inverse and forward enabled:
\begin{itemize}
    \item Model training is solve the inverse problem: $\theta = F^{-1}(d)$
    \item Model inference is solve the forward problem: $d = F(\theta)$
    \item Backpropagation is to find $\theta$ such that $||F(\theta) - d||$ is minimized
\end{itemize}

The neural network is able to solve forward and inverse problems with one architecture, this is very remarkable. In some inverse problems, there is a concept of negative time. Currently, training data lacks explicit timestamps (representing negative time); within a given context window, the timestamp is implicitly embedded in the token's position embedding. This implicit temporal encoding aids LLM training by compensating for the absence of explicit timestamps in the content, yet it also limits the exposure of negative time issues. Since LLMs learn from text data written in the past (negative time) without explicit timestamps, the representation of the time dimension within the model is often unclear and ambiguous.

In an example from Shi et al. (2023) \cite{hallucinate2023}, such as 'How many World Cups has Argentina won?', the correct answer is three, as Argentina won in 1978, 1986, and 2022. However, if the majority of the training data for an LLM predates 2022, the model might incorrectly answer with two. The reason is that the training data is weighted equally before and after 2022. By incorporating negative time, explicit content timestamps, and embedding these into the model similar to position embeddings, there is a high likelihood of obtaining 'three' as an answer.

\subsection{Dynamic computation with infinite degree of freedom}
\label{dynamic_computation}
According to the discretization theorem, actual computation is carried out at each node. We can extend the inverse problem formulation as follows

\begin{equation}
    \theta = F^{-1}(d) \approx \sum_{n=1}^{\text{nodes}} f_n(x_n)
\end{equation}

$$f_n\textnormal{ is an integrable function, } x \textnormal { is an input value.}$$

According to the Numerical Manifold Method, there is a cover (cover topology) per node, which we refer to as the node cover. The node cover changes its orientation after the weight update during the backward pass. Consequently, the neural network now has an almost infinite degree of freedom, as the orientation can take on nearly unlimited directions. 

\begin{figure}[H]
    \centering
    \includegraphics[width=1\linewidth]{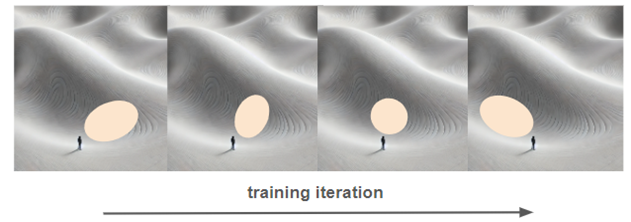}
    \caption{Dynamic Node Cover}
    \label{fig:fig4}
\end{figure}

\subsection{Exponential computation}

Neural networks can be viewed as a composition of functions. For a single pathway connecting one node per layer across all hidden layers (H) end to end, the composition of functions can be represented as follows.

\begin{equation}
    g(x) = (f_1 \circ f_2 \circ \ldots \circ f_n)(x) = f_1(f_2(\ldots f_n(x) \ldots))
\end{equation}

This notation illustrates how the output of each function becomes the input to the next, forming a nested composition of functions from the input layer to the final output layer. In the Numerical Manifold Method, a dual pairing (dual topology) connects node covers together.

\begin{figure}[H]
    \centering
    \includegraphics[width=0.75\linewidth]{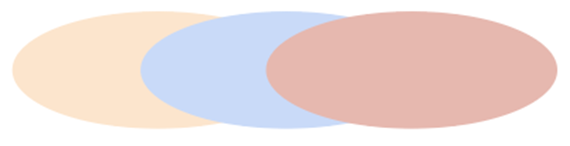}
    \caption{Dual Pairing}
    \label{fig:fig5}
\end{figure}

Then we can extend composition of functions as following

\begin{equation}
    f_1(f_2(\ldots f_n(x) \ldots)) \approx \sum_{h=1}^{H} f(x^h)
\end{equation}

A fully connected neural network, such as a Multi-Layer Perception (MLP), can be defined by its hidden layers and the nodes within those hidden layers.

\begin{equation}
    \sum_{n=1}^{\text{nodes}} f_n(x_n) \rightarrow \sum_{h=1}^{H} \sum_{w=1}^{W} f(x^h)
\end{equation}

Where:
\begin{itemize}
    \item $H$ is the total number of hidden layers.
    \item $W$ is the total number of node per hidden layer.
    \item $x$ is input value on node $[h,w]$.
    \item $f$ is a linear polynomial function at node $[h,w]$, like the ReLU activation function.
\end{itemize}

The total exponential computation power can be estimated by $x^H$ and is defined as the \textbf{learning capacity}. This exponential computation, along with the dynamic computation described in Section \ref{inverse_forward}, makes neural networks extremely powerful. 
Learning Capacity:  $T \to O(\lambda^{WH})$

\begin{figure}[H]
    \centering
    \includegraphics[width=0.5\linewidth]{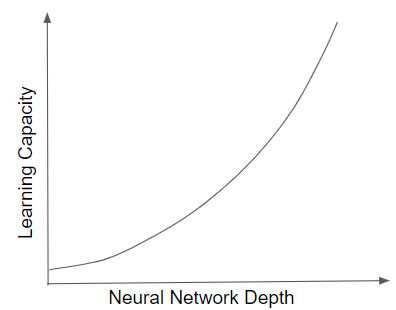}
    \caption{Exponential Computation}
    \label{fig:fig6}
\end{figure}

\subsection{Self-progressing boundary condition}
\label{self_progressing}
For any numerical computation, boundary conditions are required. Let's examine the loss function: $L(\hat{y},y)$, $\hat{y}$ is the computed output (predicted value) at the end of the forward pass, while $y$ is the actual target value applied at the beginning of the backward pass. The annotation, which is the target value ($y$), becomes the boundary condition for neural networks.  This target value ($y$) is back-propagated through the neural network in a self-progressing manner. This self-progressing nature of boundary condition implementation makes self-supervised learning possible, as seen in models like GPT.

\begin{figure}[H]
    \centering
    \includegraphics[width=0.6\linewidth]{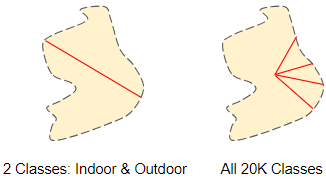}
    \caption{Learning Space: ImageNet Examples}
    \label{fig:fig7}
\end{figure}

For example, with ImageNet, we have two tasks: 1) classifying images into two categories—indoor or outdoor, and 2) classifying images into all 20K categories. While we use the same dataset (the same data manifold), the annotations differ between tasks. As a result, the learning space for the neural networks also differs. Classifying the two categories is likely much easier and can be handled by a linear, shallow network, whereas classifying 20K categories requires a deeper, more complex or high order nonlinear network.

\textbf{Learning space}($\mathbb{R}^{LS}$) is defined as a data space ($\mathbb{R}$) or data manifold after annotation ($\Omega$).
\begin{equation}
    \mathbb{R}^{LS} \leftarrow \Omega\mathbb{R}, \Omega \in \textnormal{annotation}
\end{equation}

\begin{figure}[H]
    \centering
    \includegraphics[width=0.75\linewidth]{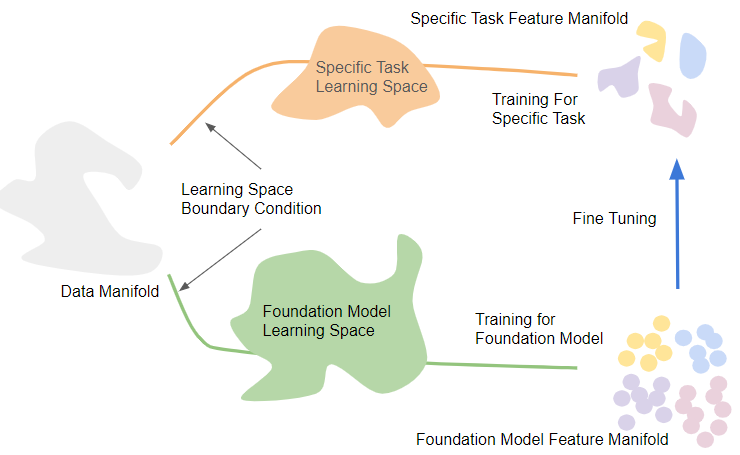}
    \caption{Learning Space Overview }
    \label{fig:fig8}
\end{figure}

We define this as the learning complexity ($N$) of the learning space, which determines the requirements for the neural network's computational capacity.
\begin{equation}
    N \to O(\mathbb{R}^{LS\cdot J})
\end{equation}

We explicitly highlight the degree of non-linearity ($J$) because we believe it is a determining factor in the learning space. 

\subsection{Deep Manifold}
\label{deep_manifold}
Based on the principles of the Numerical Manifold Method, we name each node as a node cover, and one connection between two nodes as a dual pairing. These dual pairings form a covering space (topology) connecting all nodes, which we call the \textbf{Deep Manifold}.

\textbf{Deep Manifold Space} ($\mathbb{R}^{dm}$) is defined as a space representing weight on every node of the entire neural network. Deep Manifold produces many deep manifold space during the training as described below

\begin{figure}[H]
    \centering
    \includegraphics[width=0.9\linewidth]{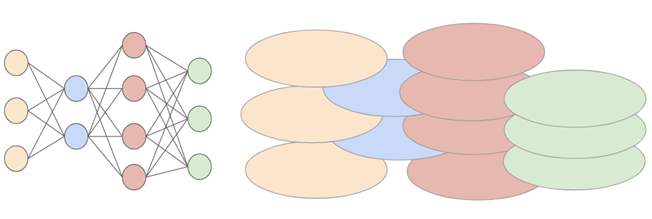}
    \caption{Deep Manifold – Neural Network Manifold}
    \label{fig:fig9}
\end{figure}

\begin{itemize}
    \item Deep Manifold Base
    \begin{itemize}
        \item Begin of training, all weights initialized
    \end{itemize}
    \item Deep Manifold Intermediate
    \begin{itemize}
        \item During the training at the end of each iteration
    \end{itemize}
    \item Deep Manifold Final
    \begin{itemize}
        \item End of training, becomes feature manifold
    \end{itemize}
\end{itemize}

\begin{figure}[H]
    \centering
    \includegraphics[width=0.9\linewidth]{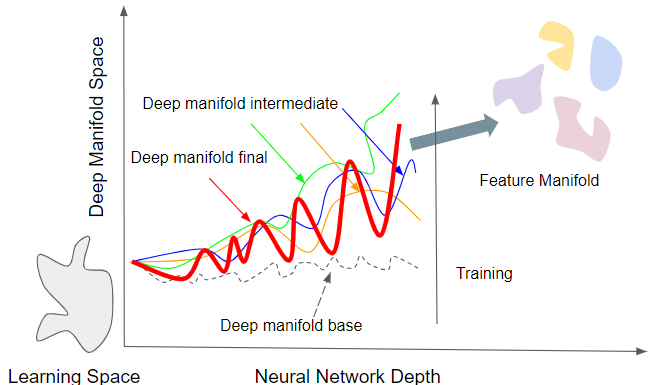}
    \caption{Deep Manifold Space}
    \label{fig:fig10}
\end{figure}

To include non-linearity explicitly in deep manifold space: $\mathbb{R}^{dm\cdot J}$

With Deep Manifold Space, we would like to highlight the following:

\begin{itemize}
    \item For a deep manifold space with high-order non-linearity, we should mathematically expect Fourier features (characteristics) within the space. The explicit manifestation of these Fourier features depends on the rigidity of the deep manifold space and the resilience of the neural network (see preceding section \ref{intrinsic_pathway})
    \item Most model interpretability efforts focus on the last or last few layers of the deep manifold space. However, the entire deep manifold space is a collection of features from all nodes and layers. While the last few layers may contain predominant features, they are not comprehensive.
\end{itemize}

\subsection{Learning Transformation}
\label{learning_transformation}
Neural network training is learning transformation powered by Learning capacity ($T$):

\begin{equation}
    T \to O(\lambda^{WH}):\mathbb{R}^{LS\cdot J} \to \mathbb{R}^{dm\cdot J}
\end{equation}

\begin{figure}[H]
    \centering
    \includegraphics[width=0.8\linewidth]{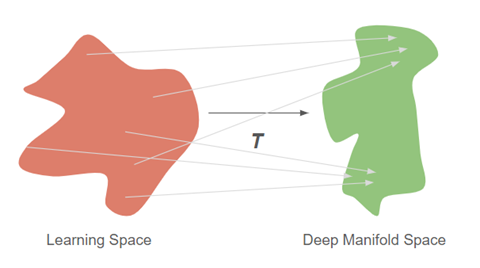}
    \caption{Learning Transformation}
    \label{fig:fig11}
\end{figure}

This means a neural network is capable of handling non-linearity. We explicitly highlight the degree of non-linearity for three reasons:
\begin{itemize}
    \item The function high order (non-linearity) represents computation power mathematically.
    \item High-order non-linearity is often the most challenging aspect of numerical computation.
    \item The non-linearity of neural networks and learning spaces is not yet well understood.
\end{itemize}

\begin{figure}[H]
    \centering
    \includegraphics[width=0.50\linewidth]{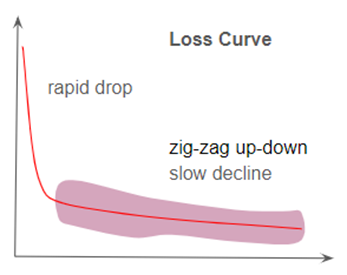}
    \caption{Loss Curve}
    \label{fig:fig12}
\end{figure}

When $J$ of  $\mathbb{R}^{dm\cdot J}$ approaches $H$ of $O(\lambda^{WH}$) , neural network starts losing learning power. This explains the zig-zag pattern observed in the loss curve during the slow decline stage of almost all foundation model training, suggesting that training is struggling to converge—a \textbf{hidden bottleneck} identified in our analysis. A model is considered to have converged effectively when the standard deviation (SD) of its loss (error) values stabilizes at less than 5. 

According to our analysis, hidden bottleneck mitigation are already incorporated into deep learning practices. They are

\begin{enumerate}
    \item Dropouts and shortcut/skip connections
    \item Data engineering work including labeling function \cite{snorkel2017}
    \item Transformer has the best mitigation strategy
\end{enumerate}

These mitigation either delay the development of non-linearity in the deep manifold space (\#1) or reduce the learning space non-linearity from the beginning (\#2). The Transformer architecture (\#3) employs the most aggressive and effective mitigation strategy: reducing non-linearity at each node in every iteration. The attention layer functions as the data processing layer, effectively reducing non-linearity. In addition to the data processing concept, the Transformer includes two implementations that align with the principles of the numerical manifold method.

As the strength of the bottleneck increases, so does the rigidity of the deep manifold space. At this point, Fourier features within the space should become apparent. Fourier analysis helps in understanding the frequency components that constitute the deep manifold space response. In general, the more rigid a system is, the more it tends to exhibit high-frequency components in its response. In this context, Fourier analysis can be an effective method for measuring learning capacity during training.

We are surprised by the prolonged slow decline, which is accompanied by a relatively high standard deviation (SD) in the loss values. This suggests that other factors may be at play.

\begin{figure}[H]
    \centering
    \includegraphics[width=1.0\linewidth]{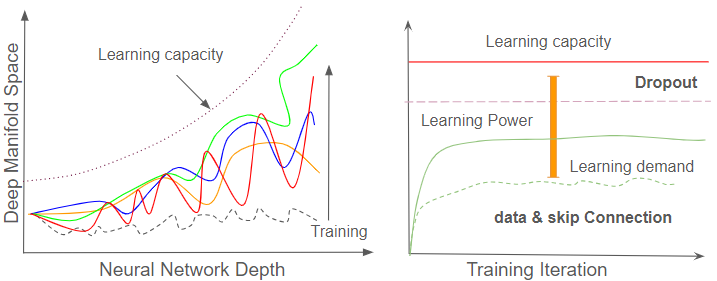}
    \caption{Neural Network Hidden Bottleneck}
    \label{fig:fig13}
\end{figure}

\subsection{Neural Network Intrinsic Pathway}
\label{intrinsic_pathway}
The prolonged, slow decline in the loss curve suggests that the neural network is a robust and resilient system. We have concluded the following from the previous sections
\begin{enumerate}
    \item Dynamic computation with infinite degree of freedom (section \ref{dynamic_computation}).
    \item The fluidity in self-progressing boundary conditions (section \ref{self_progressing}).
\end{enumerate}

The neural network operates as a power-efficient system, with each node requiring minimal computational power, even when the deep manifold space becomes rigid and a bottleneck develops. Additionally, all foundation model pre-training is self-supervised. The neural network's self-progressing boundary condition imposes no restrictions on where incoming data is processed. Incoming data will be directed to whichever nodes are capable of processing it. This means that the \textit{neural network continues to learn even during the slow decline stage}. In this sense, grokking and double descent are evidence of this continued learning.

It also means that the same token will be processed in different nodes. It is highly likely that many replicas of identical or near-identical \textbf{feature bits} (units of feature) disperse throughout the network. The inequality in mathematics, as described in the 'Contact Theory' (Shi, G. 2015) \cite{Shi2015-ml}, suggests that connections between nodes (pathways) are not equal due to random neural network initialization. Our working theory proposes that feature bits propagate through the network, with their propagation distance determined by the computational capacity of each node. The pathway appears to be power-driven, prioritizing certain features or patterns during learning in a discriminatory manner. While this \textbf{Intrinsic Pathway} (IP) is mathematically plausible, the underlying theory remains unclear. It seems that neural networks are leading us into the realm of bifurcation theory.

\begin{figure}[H]
    \centering
    \includegraphics[width=1.0\linewidth]{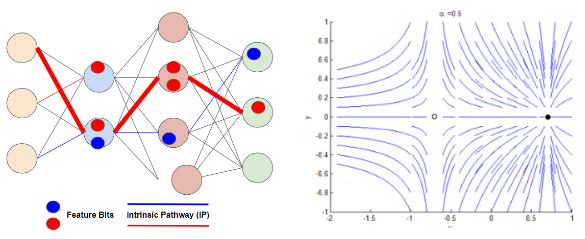}
    \caption{Feature Bits \& IP (left) and Bifurcation theory illustration(right)}
    \label{fig:fig14}
\end{figure}

It prompts a fundamental question regarding how to measure the completeness of training. This could have numerous implications for pre-training and post-training, such as fine-tuning, in-context learning, model compression and merge.

\subsection{Neural Network Fixed Point and Convergence}
\label{fixed_point}
A fixed-point theorem in mathematics states that a function F will have at least one point x such that F(x) = x provided certain general conditions are met for the function F. The general condition is learning space (section \ref{self_progressing}) for neural networks. The theory of Fixed Point Classes \footnote{Gen-Hua Shi contributed about 1/3 of the content of this book. Prof.Kiang was his advisor in the 1960s} (Kiang Tsai-Han, 1980) \cite{kiang1980}, a concept rooted in algebraic topology and fixed-point theory, suggests the existence of numerous fixed points. Yet, not all these fixed points necessarily represent true or final solutions. We establish a linkage between this false fixed point and the occurrence of hallucinations in foundation models. This linkage offers a promising pathway for further research into the mechanisms underlying hallucinations. The hallucinations have many contributing factors or sources. One of them is the convergence due to inadequate learning space (section \ref{self_progressing}) and hidden bottleneck (\ref{learning_transformation}). 

\begin{figure}[H]
    \centering
    \includegraphics[width=0.6\linewidth]{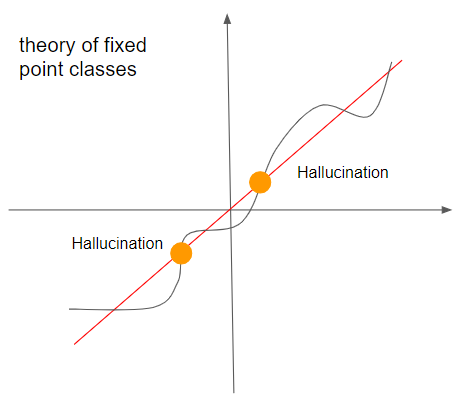}
    \caption{Neural Network Fixed Point}
    \label{fig:fig15}
\end{figure}

\section{Discussion}
\label{discussion}
We found that neural networks function as a numerical computation framework, and more precisely, they operate as a form of numerical manifold. The question then arises: if they aren't numerical computation, what kind of computation are they doing? For numerical computation, the primary focus is on convergence—both its reliability and speed.

\subsection{Convergence and Training Completion}
\label{convergence_1}
In most physics-based numerical computations, equilibrium states, either static or dynamic, are typically derived from theoretical or empirical formulations and can be viewed as fixed points. In contrast, neural networks may not exhibit such fixed points. Instead, they likely operate around the center of a Gaussian-symmetric manifold, forming an equilibrium distribution. This can resemble the cyclic pattern observed in the 'first two principal components of the learned input embedding,' as reported by Liu et al. (2022) \cite{grokking2022}. Drawing inspiration from the concept of level curves (Sayama, 2015)\cite{sayama2015}, we envision a Gaussian-symmetric manifold composed of interconnected Toruses. 

\begin{figure}[H]
    \centering
    \includegraphics[width=0.7\linewidth]{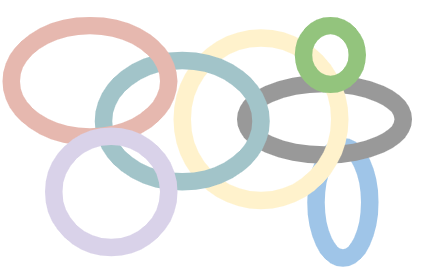}
    \caption{Interconnected Toruses}
    \label{fig:fig16}
\end{figure}

This brings into question the current criterion for training completion, which relies on the loss value relative to the annotations. However, annotations themselves are not convergence points or fixed points; rather, they serve as boundary conditions (Section \ref{self_progressing}) that guide the direction of convergence without defining the convergence point. How can we define training completion without knowing convergence point(s) ? 

\subsection{Convergence and Boundary Condition}
\label{convergence_2}
Boundary conditions play a crucial role in numerical computation. As outlined in Section \ref{self_progressing} (Learning Space), the boundary condition in a neural network corresponds to annotations, which are embedded within the training data for supervised or self-supervised learning, shaping the learning space. These boundary conditions can be categorized into two aspects: 1) weak annotations (labels) with noise, and 2) symmetric boundary conditions. Similar to other numerical computations, boundary conditions guide the direction or path toward convergence.

Weak annotations are easier to obtain than strong annotations (ground truth), as demonstrated by the Snorkel approach (Ratner et al., 2017) \cite{snorkel2017}.  Multiple weak annotations, such as using three weak labels, can provide a more reliable direction for convergence compared to relying on a single strong annotation. Additionally, since training data is fed iterative in batches, neural network training can be viewed as a dynamic process. According to perturbation theory, noise plays a beneficial role in aiding convergence. In fact, noise within weak annotations introduces small perturbations into the training process, which can enhance feature refinement, like feature purification noted by Allen-Zhu and Li (2022) \cite{allen-zhu2022}.

Symmetric boundary conditions have two key effects. First, oppositely oriented noise tends to cancel out, reinforcing the benefit of weak annotations as discussed above. Second, symmetric boundary conditions enhance both the stability and speed of convergence, providing further evidence for the effectiveness of contrastive learning and the Maximum Manifold Capacity Representations (MMCRs) framework proposed by Yerxa et al. (2023) \cite{mmcr2023}. 

The neural network boundary condition is self-progressing and data type agnostic, not only supports multimodal natively, but also has a huge advantage over physical/biological based numerical computation. FourCastNet (Kurth et al., 2024) \cite{fourcastnet2022}  serves as an excellent example: \textit{‘The Earth system modeling enterprise, including weather forecasting and climate predictions, generates tens of petabytes of data every year and that quantity is growing exponentially. Yet, the gigantic volume of data and wealth of information contained therein is often collapsed into a few simple metrics that guide decision-making’}. Traditional physics-based numerical weather prediction models typically use fewer than 50 variables and rely on supercomputers to perform extensive computations for convergence. FourCastNet leverages the self-progressing and data type agnostic of neural networks boundary conditions, enabling the integration of more diverse data sources, such as weather satellite images, into the model. This provides additional convergence guidance and significantly accelerates the convergence process. As a result, FourCastNet is 80,000 times faster in predictions and 10,000 times more energy-efficient than the Integrated Forecasting System.
\subsection{Convergence and High Order Non-linearity}
\label{convergence_3}
High-order non-linearity becomes increasingly critical and may present a bottleneck for foundation models processing trillions of tokens. On the surface, techniques like Chain of Thought (CoT) exemplify high-order non-linearity with their complex, multi-step reasoning. Within the transformer architecture, normalization plays a pervasive role, occurring throughout the entire network. The reason for normalization is to prevent the divergence. 

Convergence in high-order nonlinear systems is particularly challenging, as even small errors can quickly lead to divergence. The key lies in balancing the preservation of high-order nonlinear features (information) while minimizing the computational complexity associated with them. From an architectural standpoint, techniques like dropout and skip connections effectively mitigate the degree of high-order non-linearity. From the data perspective, weak annotations (with noise) and symmetric distributions help reduce the computational demands of handling high-order non-linearity. 

There are a good number of alternatives to gradient descent for achieving convergence. However, the current implementation of gradient descent presents a limitation in the training of foundation models. This issue isn't exclusive to foundation model training. When high-order non-linearity is not accurately modeled or computed, immense computational power—such as that provided by supercomputers—is required, as seen in tasks like 3D seismic discontinuity analysis and geophysical modeling.

\section{Future Work}
This Part 1 work aims to establish key baselines grounded in solid mathematical reasoning. It should be viewed as a conceptual and directional study. Further rigorous proofs and empirical validation are necessary to fully substantiate the merits of the ideas presented in this paper.

\section{Acknowledgments}
Our initial motivation was to understand the high-order non-linearity of neural networks. The first author has been suspecting the presence of high-order non-linearity in neural networks since 2017. It was Prof. James Zou's talk on YouTube that inspired the first author to persist in pursuing a deeper understanding of this high-order non-linearity, despite the apparent lack of interest from others. Deep Manifold emerged as a by-product of this investigation into high-order non-linearity and was discovered by accident. 

\bibliographystyle{splncs04}
\bibliography{biblio}
\end{document}